\documentclass[conference]{IEEEtran}
\usepackage{times}

\usepackage[numbers]{natbib}
\usepackage{multicol}
\usepackage[bookmarks=true]{hyperref}
\usepackage[ampersand]{easylist}
\usepackage{siunitx} 
\usepackage{graphicx}
\usepackage[svgnames]{xcolor}
\usepackage{acro}
\usepackage{booktabs}
\usepackage[table]{xcolor}
\usepackage{amsmath, amssymb}
\usepackage{mathtools}
\usepackage{cleveref}
\usepackage{tikz}

\usetikzlibrary{arrows.meta, positioning}
\usepackage{adjustbox}
\usepackage{forest}

\DeclareAcronym{VIO}{
   short = VIO,
   long  = Visual-Inertial Odometry
 }
 \DeclareAcronym{GMM}{
   short = GMM,
   long  = Gaussian Mixture Model
 }
 \DeclareAcronym{BNP}{
   short = BNP,
   long  = Bayesian nonparametric
 }
 \DeclareAcronym{IMU}{
   short = IMU,
   long  = Inertial Measurement Unit
 }
 \DeclareAcronym{IQR}{
   short = IQR,
   long  = interquartile range
 }
 \DeclareAcronym{PSD}{
   short = PSD,
   long  = power spectral density
 }

\begin{document}

\title{When Your State Estimator Has Lost The Plot:\newline Detecting Estimator Failures Via Spectral Analysis}

\author{\authorblockN{Christian Lanegger$^{1}$, Helen Oleynikova$^{2}$, Roland Siegwart$^{1}$, and Michael Pantic$^{1}$}\authorblockA{$^{1}$Autonomous Systems Lab, ETH Zurich, Switzerland, $^{2}$Exclaim Robotics, Zurich, Switzerland}}

\maketitle

\begin{abstract}
Reliable onboard state estimation is essential for safe robotic operation, yet unmodeled disturbances, such as sensor aliasing or out-of-distribution noise, still cause estimators to degrade or fail completely. While many methods aim to improve estimator robustness, only a few provide introspective mechanisms to assess estimate quality. Existing uncertainty measures, such as covariances, rely on idealized assumptions and tend to be overconfident, and more recent data-driven approaches are typically tied to their training data distributions.

We propose a sensor-agnostic introspective method that assesses estimator health by analyzing the frequency-domain power distribution of recent velocity estimates. The method is evaluated using outdoor flight data from an aerial robot running visual–inertial, LiDAR–inertial, and radar–inertial odometry. The dataset includes multiple estimator failures, enabling analysis of several frequency-domain indicators, such as signal power, spectral bandwidth, and entropy. We observe consistent spectral power differences between healthy and degraded estimates, allowing detection of \SI{51}{\%}–\SI{58}{\%} of labeled failures with \SI{60}{\%}–\SI{84}{\%} precision across three fundamentally different state estimation frameworks. Our results show that even a simple frequency-domain analysis of a state estimator's output can serve as a lightweight introspective tool to complement existing robustness techniques in real-world robotic deployments, and opens promising avenues for future investigation.
\end{abstract}

\IEEEpeerreviewmaketitle

\section{Introduction}
When sitting in a stationary train that is about to depart, briefly observing a neighboring train begin to move can create a moment of confusion. Although no accelerating forces are felt, visual cues suggest motion, and the assumption of a static environment implies that one must be moving. This disagreement between sensory inputs pushes us to further inspect the surroundings until the inconsistency is resolved and it becomes clear that the other train is in motion. This ability to reason about conflicting sensory information and reassess the reliability of one’s perception is a form of introspection that humans rely on to handle unforeseen disturbances.

Robotic systems, in contrast, often lack such introspective capabilities. Yet, as robots are increasingly deployed outside controlled environments, the ability to assess the reliability of their own state estimates becomes critical. This challenge is particularly pronounced for aerial robots, which are gradually being deployed for tasks such as inspection of bridges, wind turbines, and other critical infrastructure. These applications rely not only on high pose accuracy but also on robustness. When operating in close proximity to structures, even small errors in the state estimate can result in collisions with the environment and fatally damage the robot. Additionally, for such tasks aerial robots must operate under highly variable conditions. Illumination, visual and geometric features, available free space, and GNSS visibility can change drastically over time and across environments. No single sensing modality operates reliably across all such conditions, making multi-sensor state estimation essential, but also more susceptible to sensor degradation and unmodeled disturbances.
\begin{figure}[t]
\centering
\includegraphics[width=\columnwidth]{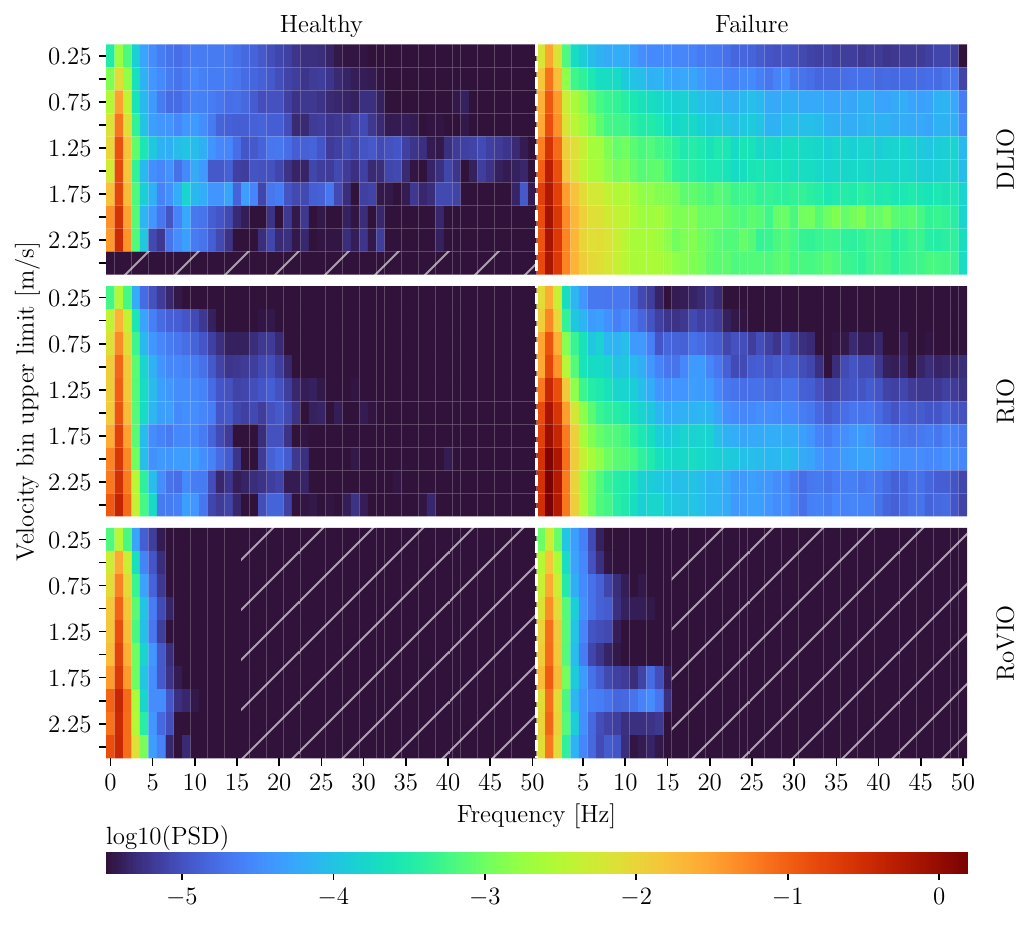}
\caption{Power spectral density (PSD) over the 0–\SI{50}{\hertz} frequency range for three different state estimators. The PSD is shown for healthy and failure-labeled data across different velocities, revealing clear differences in power distribution for faulty estimates for all estimators. DLIO contains no healthy data above \SI[per-mode=symbol]{2.25}{\metre\per\second} in the dataset, and RoVIO is limited to a maximum PSD resolution of \SI{15}{\hertz}; corresponding values are crossed out.}
\label{fig:heatmap_mean_psd}
\end{figure}

Deploying complex mobile robots outside controlled laboratory environments therefore requires state estimators that fuse heterogeneous sensing modalities and, more importantly, can handle sensor failures and measurement degeneracies. Existing approaches mainly address robustness to Gaussian noise or expected degeneracies, yet they often remain vulnerable to non-Gaussian or unmodeled disturbances. Leveraging sensor diversity and redundancy—while avoiding single points of failure—is essential to maintain functionality under such conditions \cite{prorok2021taxonomy}. Likewise, decomposing the state-estimation pipeline into multiple sub-problems has been shown to increase resilience and preserve functionality during sensor dropouts \cite{lanegger2023chasing, nubert2022graph}. Maintaining estimator performance in the presence of unexpected disturbances requires detecting them. While dropouts are straightforward to identify, unmodeled disruptions require introspective methods capable of quantifying estimator trustworthiness with minimal prior assumptions about failure causes. To date, only a few works have directly addressed this problem \cite{raanan2018detection, lanegger2023fuse} and are robot-specific or scale badly with the number of sensors, leaving a clear gap in the literature.
Using datasets obtained on an aerial robot running multiple onboard estimators (Lidar-, \mbox{Visual-,} and Radar-Inertial frameworks) which experienced various failure modes during flight, we investigated how the output signals of different state estimators behave in healthy and degraded/faulty states. Human operators are often able to tell that ``something'' is about to go wrong from time series and plots, without knowing internal states of the algorithms. How can we formalize such a signal and what is the working mechanism behind it?
\subsection{Contribution}
In this work, we show that spectral analysis can be used as an introspective tool to detect state estimator degeneracies in any kind of IMU-sensor fusion, regardless of modality (vision, lidar, radar) or algorithm (filtering or optimization). 
Compared to other methods, our contributed method does not rely on multi-sensor consensus and allows us to model the estimator as a black box, which addresses an often overlooked gap in state estimation literature. We present the ideation, design, and application of the method using labeled multi-sensor data obtained in the wild. 
Our analysis shows that many failures are reflected in the spectral characteristics of the velocity estimate, particularly through an increase in power in the higher-frequency bands. This observation forms the basis of the proposed state-estimator-integrity test which requires only (i) a velocity estimate output at a sufficiently high enough frequency and (ii) fusion of at least one sensor with an IMU. Our method specifically complements existing literature in case of unmodeled noise and unknown disturbances: a crucial capability for real-world robot deployment.
\section{Related Work}

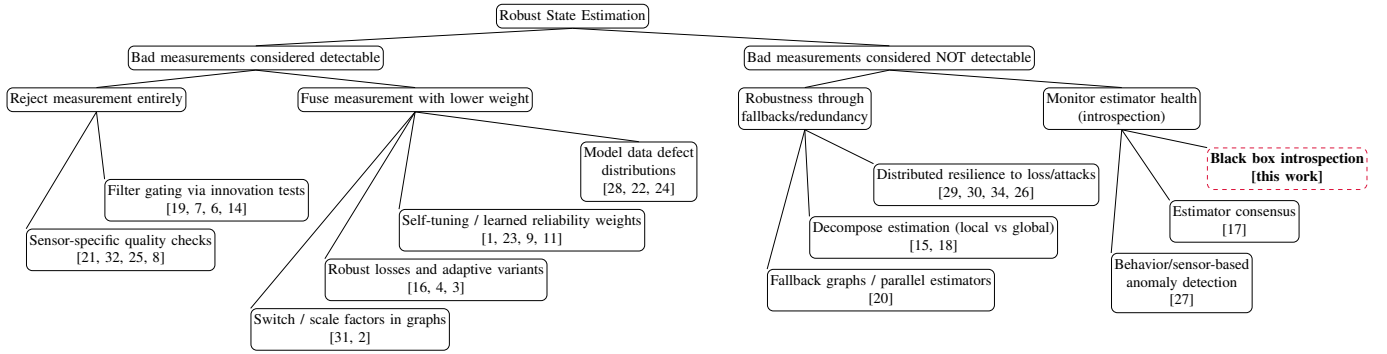
\begin{figure*}
\resizebox{\textwidth}{!}{
\begin{forest}
  for tree={
    draw, rounded corners,
    grow=south,
    l=0,
    s sep=10pt,
    align=center,
    inner sep=2pt,
    parent anchor=south,
    child anchor=north
  }
 [Robust State Estimation
    [Bad measurements considered detectable
        [Reject measurement entirely, grow=-60
            [Sensor-specific quality checks\\\cite{nubert2022learning,tuna2023x,pfreundschuh2024coin,Brunner2014}, child anchor=north west]
            [Filter gating via innovation tests\\\cite{lynen2013robust,brumback1987chi,brommer2020mars,hausman2016self}, child anchor=north west]
        ]
        [Fuse measurement with lower weight, grow=-60
            [Switch / scale factors in graphs\\\cite{sunderhauf2012switchable,agarwal2013robust}, child anchor=north west]
            [Robust losses and adaptive variants\\\cite{huber1992robust,black1996robust,barron2019adaptive}, child anchor=north west]
            [Self-tuning / learned reliability weights\\\cite{agamennoni2015self,pfeifer2017dynamic,chen2019selective,choi2025statistical}, child anchor=north west]
            [Model data defect\\distributions\\\cite{rosen2013robust,olson2013inference,pfeifer2021advancing}]
        ]
    ]
    [Bad measurements considered NOT detectable
        [Robustness through\\fallbacks/redundancy, grow=-45,before computing xy={s=-2cm}
            [Fallback graphs / parallel estimators\\\cite{nubert2022graph}, child anchor=north west]
            [Decompose estimation (local vs global)\\\cite{he2025ligo,lanegger2023chasing}, child anchor=north west]
            [Distributed resilience to loss/attacks\\\cite{saldana2017resilient,schlotfeldt2021resilient,wheeler2019switching,prorok2021taxonomy}, child anchor=north west]
        ]
        [Monitor estimator health\\(introspection), grow=-45,before computing xy={s=5.5cm}
            [Behavior/sensor-based \\anomaly detection\\\cite{raanan2018detection}, child anchor=north west]
            [Estimator consensus\\\cite{lanegger2023fuse}, child anchor=north west]
            [Black box introspection\\ {[}this work{]}, draw=Crimson, dashed, child anchor=north west,font=\bf]
        ]
    ]
]
\end{forest}
}
\caption{Overview of related literature, structured by the assumption of detectability of bad measurements.}
\label{fig:robustness_mechanisms_rss_fullwidth}
\end{figure*}

Most modern on-board state estimators rely on the fusion of multiple sensors with complementary properties. Regardless of the sensor modalities, there are always defects, such as (non-gaussian) noise, degradation, outliers, and aliasing. The choice of tools used to robustify a state estimator against data defects hinges on the detectability of bad measurements. \Cref{fig:robustness_mechanisms_rss_fullwidth} provides a detailed and structured overview of the related literature with citations. The rest of this chapter summarizes the coarse structure of the literature and details some illustrative examples.
If the quality of measurements can be determined fairly reliably, bad measurements can either be discarded or fused into the estimate with weights/covariances that take the quality into account.
If the detection of bad measurements is considered impossible or too unreliable (\textit{a priori} or \textit{a posteriori}), we cannot discard or down-weight such measurements. Including such measurement results in the slow degradation of the state estimate, leading to either divergence of state estimator over-confidence over time. Here, robustness is often achieved through fallbacks or redundant estimators and external monitoring of estimator health. 
In general, we consider active monitoring of estimator health through introspection to be underdeveloped in the literature. 
While there is previous work that uses consensus between estimators~\cite{lanegger2023fuse} or anomaly detection~\cite{raanan2018detection}, there is a gap in the literature when it comes to inferring estimator health based only on the qualities of their output (except simple covariance/out-of-bounds thresholds).
In the following, relevant examples of the different methods are given and discussed.
\subsection{Measurement rejection}
Measurement rejection methods are often tied to a specific sensor modality. For example, LiDAR-based frameworks can exploit the geometric properties of scans to assess whether the measurements sufficiently constrain the state estimate \cite{nubert2022learning, tuna2023x}. Vision-based systems recently greatly benefited from data-driven methods, for example, for training end-to-end \ac{VIO} systems robust to corrupted or missing data \cite{chen2019selective}, or for down-weighting unreliable visual features using self-supervised multi-view consistency \cite{choi2025statistical}.

\subsection{Measurement down-weighting}
Backend robustness techniques aim to mitigate outliers at the estimation level. Filter-based approaches commonly rely on Mahalanobis distance thresholding \cite{lynen2013robust} or statistical validation via $\chi^2$ tests \cite{brumback1987chi, brommer2020mars, hausman2016self}. In graph-based optimization, Switchable Constraints \cite{sunderhauf2012switchable} and Dynamic Covariance Scaling \cite{agarwal2013robust} down-weight inconsistent measurements and are functionally equivalent to robust M-estimators. M-estimators down-weight residuals that do not fit a Gaussian distribution at the cost of introducing at least one additional tuning parameter, which can be self-tuned using expectation maximization \cite{agamennoni2015self} or directly estimated by modeling sensor variance as a state variable \cite{pfeifer2017dynamic}. All these methods assume Gaussian noise, an assumption frequently violated in practice. Non-Gaussian alternatives such as \acp{GMM} have been proposed \cite{rosen2013robust, olson2013inference}, but often suffer from convergence issues, partially addressed in later work \cite{pfeifer2021advancing}.

\subsection{Redundancy and fallbacks}
If bad measurements cannot be detected individually, the estimator architecture can be made redundant by running multiple estimators that act as fallbacks. 
Avoiding single points of failure by exploiting redundancy and diversity is central to resilience \cite{prorok2021taxonomy}, and has recently been adopted in sensor fusion. Examples include fallback factor graphs for GNSS outages \cite{nubert2022graph} and decomposed estimation pipelines that improve robustness and accuracy \cite{he2025ligo, lanegger2023chasing}.

\subsection{Health monitoring}
Despite these advances, most approaches target known disturbances. Truly resilient systems must also handle unknown and unmodeled failures, motivating the development of introspective methods with minimal assumptions. Existing health-monitoring techniques are limited: cross-comparing estimator outputs through consensus scales poorly and requires multiple reliable estimators \cite{lanegger2023fuse}, while behavior-based models such as \ac{BNP} topic modeling \cite{raanan2018detection} struggle to distinguish failures from nominal dynamics in highly agile aerial robots. Our method is what we consider an essential next step in health monitoring: an introspective failure detection method that does not rely on known noise models, modeled disturbances, or sensor consensus. 

\section{Why not just use covariances?}
\label{sec:covariances}
Covariances are the uncertainty measure that is most commonly provided by state-of-the-art state estimators. In practice, however, they are rarely used as a robust health measure, simply because they often fail to reflect the true estimate uncertainty. One key reason is that sensor models are often neither known accurately nor Gaussian. This causes the reported covariance to often underestimate the true uncertainty and lose its probabilistic meaning. As a consequence, measurement covariances are typically not set to represent true sensor uncertainty, but are instead tuned to weight the influence of measurements on the state estimate in a way that yields acceptable performance for a given robot and operating environment. While the nominal sensor model may serve as a prior for this tuning process, unforeseen disturbances remain unmodeled. This mismatch often leads to underestimated uncertainty and overly confident state estimates.

Another contributing factor is that model mismatches and unmodeled disturbances frequently manifest as changes in IMU bias estimates. Since these biases are co-estimated and not directly observable, the optimization backend may converge to a solution with small residuals and seemingly consistent covariances, even though the resulting state estimate—particularly the pose—is incorrect, essentially treating the biases as a ``dumping ground'' for uncertainty.
\begin{figure}[h]
\centering
\includegraphics[width=\linewidth]{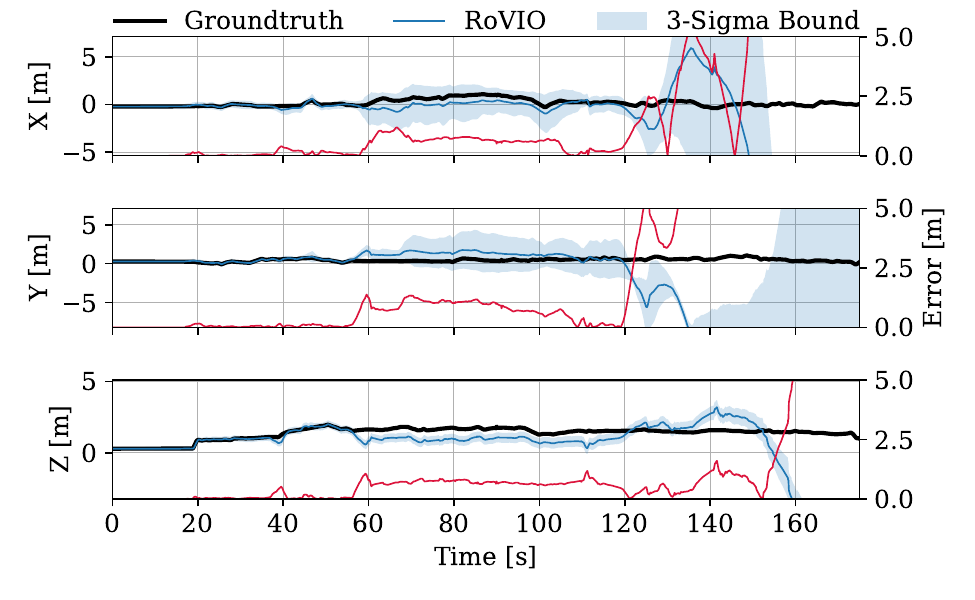}
\caption{3-Sigma bounds (in light blue) output by RoVIO showing that covariances often do not quantify the uncertainty reliably. The red curve visualizes the error between estimate and ground truth.}
\label{fig:covariances}
\end{figure}     
An example of this behavior is shown in \Cref{fig:covariances}, which compares position estimates from the VIO framework RoVIO \cite{bloesch2017iterated} with the ground truth position during flight. We intentionally induced estimator failure through fast motions that create motion blur and flying towards feature-less surfaces. While the ground truth remains largely within the $3\sigma$ bounds along the $x$-axis, this is not the case for the $y$- and $z$-axes at approximately \SI{60}{\second} and \SI{120}{\second}. Here the estimator initially drifts and subsequently diverges. We intended to perform similar tests for the LiDAR-Inertial Odometry framework DLIO~\cite{chen2022dlio}. However, DLIO does not output state covariances. This, in itself, shows that practitioners do not generally rely on covariances and highlights the need for alternative inconsistency representations that better capture and flag estimator degradation.
In \Cref{sec:methodology} we analyze how changes in power distribution across frequency bands of a state estimate can serve as such a representation.   

\section{Preliminaries}
To characterize the spectral content of a signal, we estimate its \ac{PSD} using Welch’s method \cite{welch1967fft} which reduces variance and spectral leakage while locally approximating a stationary signal. Welch's method divides the 1-D signal $x$ into $M$ overlapping segments of length $L$ with overlap $D$. Each segment is detrended, i.e., its mean is removed, and then a modified periodogram $\hat{S}_{xx}^{(m)}$ of each segment $m$ is computed
\begin{equation}
\hat{S}_{xx}^{(m)}[f_k]
=
\frac{1}{L f_s U}
\left| X_{\mathcal{B}}^{(m)}[k] \right|^2,
\end{equation}
where $X_{\mathcal{B}}^{(m)}[k]$ is the fast Fourier Transform (FFT) of the windowed segment, $f_k = k f_s/L$ is the $k$-th frequency bin, $f_s$ is the sampling frequency and $U$ is a window normalization factor.
The \ac{PSD} estimate for the full signal is obtained by averaging across all segments,
\begin{equation}
\hat{S}_{xx}[f_k] = \frac{1}{M} \sum_{m=0}^{M-1} \hat{S}_{xx}^{(m)}[f_k].
\end{equation}
The selection of segment length, overlap, and windowing function depends on the intended use of the \ac{PSD} and the spectral characteristics of interest, and is discussed further in \Cref{sec:methodology}. To evaluate the ability of \acp{PSD} to distinguish healthy from degraded estimates, we analyze several band-limited spectral metrics, including power content, spectral centroid, spectral bandwidth, and spectral entropy.
The total power contained in a frequency band $[f_1,f_2]$ is computed as
\begin{equation}
P_x(f_1,f_2)
=
\sum_{f_k=f_1}^{f_2}
\hat{S}_{xx}[f_k]\;\Delta f,
\quad
\Delta f = \frac{f_s}{N}.
\end{equation}
with $N$ being the total number of estimates used to compute the \ac{PSD}. The spectral centroid,
\begin{equation}
f_{sc}^x =
\frac{\sum_{f_k=f_1}^{f_2} f_k\,\hat{S}_{vv}^{x}[f_k]}
{\sum_{f_k=f_1}^{f_2} \hat{S}_{vv}^{x}[f_k]},
\end{equation}
indicates where most of the power is concentrated. The spectral bandwidth,
\begin{equation}
\sigma_{sb}^x =
\sqrt{
\frac{\sum_{f_k=f_1}^{f_2} (f_k - f_{sc}^x)^2\,\hat{S}_{vv}^{x}[f_k]}
{\sum_{f_k=f_1}^{f_2} \hat{S}_{vv}^{x}[f_k]}
},
\end{equation}
quantifies how spread out the power is around the centroid. The spectral entropy,
\begin{equation}
H_s^x =
- \sum_{f_k=f_1}^{f_2} p_k \log_2 p_k,
\quad
p_k =
\frac{\hat{S}_{vv}^{x}[f_k]}
{\sum_{f_k=f_1}^{f_2} \hat{S}_{vv}^{x}[f_k]},
\end{equation}
measures irregularity in the frequency distribution.
\subsection{Aerial Robot Dataset and Failure Labeling}
\label{sec:dataset}
\begin{figure}[h]
\centering
\includegraphics[width=\columnwidth]{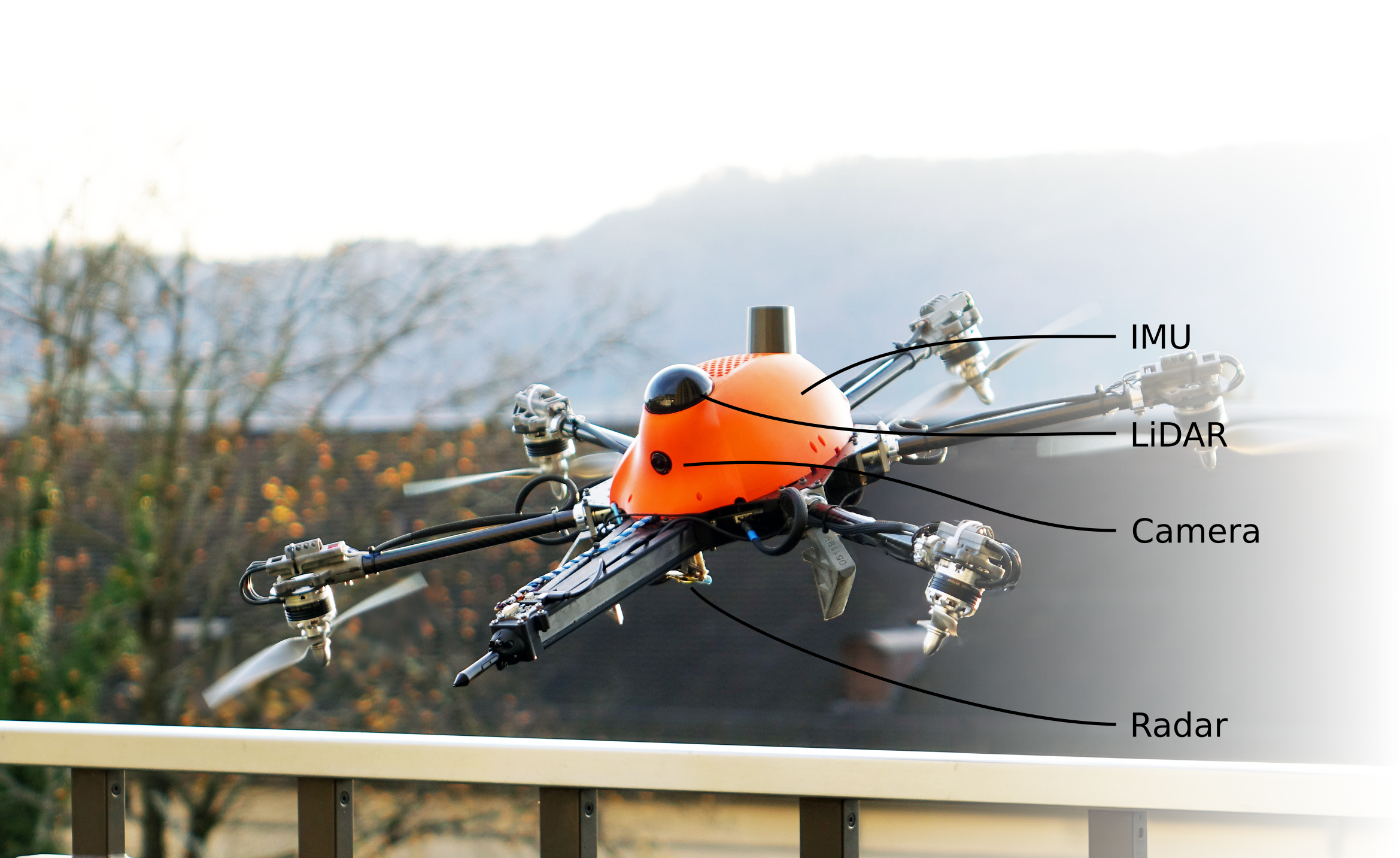}
\caption{Aerial Robot used for data collection equipped with three different exteroceptive sensors: a LiDAR, a radar, and a camera.}
\label{fig:omav_arche}
\end{figure}
We collected \SI{23}{\minute} of data from an aerial robot, shown in \Cref{fig:omav_arche}, operating outdoors to evaluate the introspective spectral metrics. The dataset contains multiple instances of estimator degradation and divergence caused by factors including changing lighting conditions, fast motions, sensor vibrations due to suboptimal mounting, limited onboard computation, and moving objects in the sensors’ fields of view.

We ran three different state estimation frameworks on the robot: RoVIO running at \SI{30}{\hertz}, DLIO producing estimates at \SI{100}{\hertz}, and an in-house graph-based sensor fusion framework built on GTSAM~\cite{gtsam} with a radar–inertial frontend based on BRIO~\cite{girod2024brio}, which outputs estimates at \SI{200}{\hertz}. We refer to this framework as RIO.

Evaluating the spectral characteristics of healthy and degraded estimator behavior requires ground-truth labeling. In outdoor environments, however, determining the exact onset of estimator failure is difficult due to the absence of reliable ground-truth pose information. To address this, we labeled failures based on consensus among the three estimators. The data was divided into \SI{0.1}{\second} windows. For each window we computed the mean absolute velocity error between all estimator pairs. If at least two estimators produced velocity estimates within a predefined error threshold, these estimates were labeled as healthy. If the third estimator exceeded the threshold relative to both others, it was labeled as faulty. The error threshold was set to \SI[per-mode=symbol]{0.2}{\metre\per\second}, corresponding to a drift of \SI{0.02}{\metre} over the window duration. This approach requires that at least two estimators remain reliable. In cases where multiple estimators exhibited clear drift or divergence, we manually labeled the affected segments. Using this procedure, we identified approximately \SI{1.5}{\minute} of degraded or diverging estimates and \SI{17}{\minute} as healthy. For data labeled as healthy, platform velocities ranged from 0 to \SI[per-mode=symbol]{2.32}{\metre\per\second}, with an average of \SI[per-mode=symbol]{0.14}{\metre\per\second}, indicating that most healthy data corresponds to low-speed motion.

Velocities were chosen for labeling rather than poses as our method quantifies estimator health based on the accuracy of velocities they are expressed in the robot’s body frame, eliminating the need for frame alignment, another potential sources of error.

\section{Spectral Analysis as tool for introspection}
\label{sec:methodology}

In this section we present and analyze the proposed frequency-domain approach, demonstrating how estimator failures manifest in the power spectral density of estimates and how these characteristics can be used to distinguish healthy from degraded outputs. Modern state estimation frameworks output body-centric estimates of pose, velocity, and \ac{IMU} biases. 
In this work, we focus only on linear velocities expressed in the body frame. Linear velocity estimates tend to be more dynamic, reacting faster to estimator degradation than pose estimates. Angular velocities are not considered, since they are typically taken directly from \ac{IMU} measurements rather than being part of the estimate.
The proposed method is based on the assumption that state estimators which combine exteroceptive and proprioceptive sensors can experience internal disagreement under degenerate conditions. This disagreement manifests as increased noise in the estimated state. As a result, the distribution of signal power across the frequency spectrum changes, with particularly noticeable differences in the higher frequency bands.

To compute an estimate of the velocity's \ac{PSD} we use Welch's method with a Hann window for smoothing. A segment overlap of $L/2$ is standard for Hann's window and also works well for our use case. We also keep the segment length $L \in [N/2, N]$, with $N$ being the total number of velocities estimated within the last \SI{1}{\second}. The resulting \ac{PSD} covers the range for the frequency band $f_k \in [0, f_s/2]$ at a frequency resolution between \SI{1}{\hertz} and \SI{2}{\hertz} at the cost of increasing spectral leakage.
The distinction in power content between reliable and unreliable estimates can be seen in \Cref{fig:heatmap_mean_psd} showing the logarithm of the average \ac{PSD} of velocity estimates for healthy and faulty segments, binned by (apparent) velocity magnitude. While both healthy and faulty estimates concentrate power below \SI{3}{\hertz}, degraded estimates consistently exhibit increased power across the spectrum and higher variability. This indicates that high-frequencies power content and its larger variance is a strong indicator of estimator degradation. Power in lower frequencies, on the other hand, is dominated by the robot's motion and directly linked to the magnitude of the velocities (notice the increased power for higher velocities).

\begin{figure*}[tb]
\centering
\includegraphics[width=\textwidth]{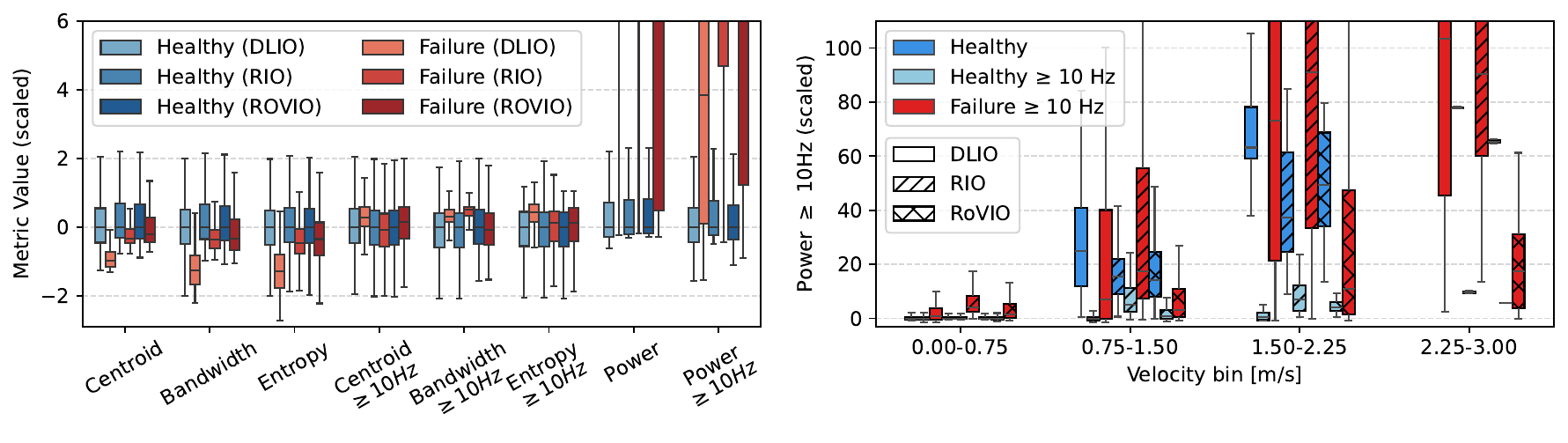}
\caption{\textit{Left:} Spectral metrics for the full frequency band and for $\geq$\SI{10}{\hertz} band of the different estimator's velocity signal during failure and trustworthy operation.\textit{Right:} Power of the full frequency and for $\geq$\SI{10}{\hertz} band for healthy and faulty distributions at different velocities. While all power increases for larger velocities, power contained in the range above \SI{10}{\hertz} increases only at failures. \textit{Note:} All values are standardized to a median and \ac{IQR} of $1$ using the healthy data of the respective spectral metric or velocity bin.}
\label{fig:metric_boxplots}
\end{figure*}

Further inspection of spectral metrics (\Cref{fig:metric_boxplots}) supports this hypothesis. Total signal power logically increases with velocity magnitude regardless of estimator health. When considering the full frequency range, the spectral centroid, bandwidth, and entropy all decrease for degraded estimates. Therefore, inspecting the full spectrum, power in the low frequencies still dominate and a good estimate of fast motion is indistinguishable from a degraded estimate. 
When excluding frequencies below \SI{10}{\hertz}, the spread of power content, quantified by its bandwidth, increases, particularly for the LiDAR- and radar-based estimators. A slight increase in spectral entropy is also observed on average, reflecting greater irregularity in the high-frequency components of degraded estimates. However, for all these metrics we observe a significant overlap in values between healthy and degraded estimates. Only absolute power content shows a clear separation. As healthy estimates do not show a significant increase in power at high frequencies, this enables the discrimination between healthy and faulty estimates.

\section{Threshold and Cut-Off Frequency Selection}

We want to find a single threshold in the high frequency power which would allow us to distinguish good from bad estimates and work across all velocity ranges. 
This threshold depends on the chosen cut-off frequency. Is there a single cut-off frequency and therefore threshold that works across all estimators and velocities? In \Cref{fig:metric_boxplots} right, we can see that at a cut-off frequency $f_{co} = 10$ \si{\hertz}, different estimators behave differently: for example, RIO has higher power in higher frequency bands than other estimators, even for healthy estimates. This indicates that a higher cut-off frequency is needed for RIO, and that the ideal cut-off is estimator-dependent. What about velocities? Ideally, the cut-off should lie just above the maximum frequency to which the robot's nominal motion contributes in terms power. As can be seen in \Cref{fig:heatmap_mean_psd}, even healthy estimates have higher power in the higher-frequency bands at higher velocities. Therefore, a higher cut-off frequency is needed for higher velocities as well, and the ideal cut-off frequency may be \textit{both} estimator- and velocity-dependent. However, a reasonable compromise might be good enough.
\begin{figure}[h]
\centering
\includegraphics[width=\linewidth]{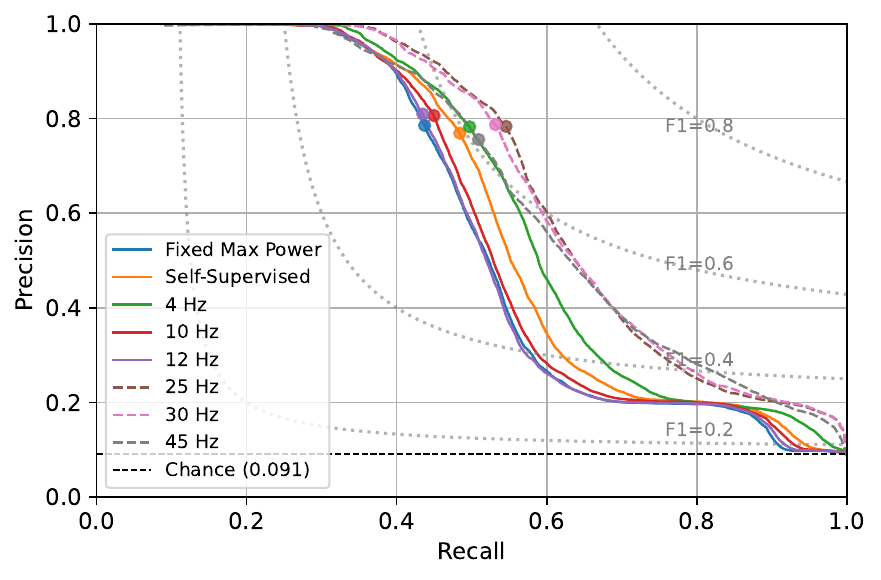}
\caption{Precision-Recall Curves for all estimators combined at different cut-off frequencies, and comparing two different methods of selecting a threshold from only healthy data (fixed max power and self-supervised). The markers label the thresholds that maximize the F1-score.}
\label{fig:pr_sensors}
\end{figure}

How can we find the optimal threshold and cut-off frequency given labeled data? \Cref{fig:pr_sensors} provides an answer: using precision-recall curves and maximizing the F1 score, it is possible to find the optimum both for individual estimators and for all estimators taken together.
When evaluated across all estimators, the best overall performance is achieved with a cut-off frequency of \SI{4}{\hertz} (\Cref{fig:pr_sensors}), detecting approximately half of all failures with a precision of \SI{78}{\%}. Evaluating the estimators individually reveals that individual higher cut-off frequencies yield improved performance: \SI{10}{\hertz} for RoVIO and \SI{25}{\hertz} for both DLIO and RIO. DLIO and RoVIO achieve precisions of \SI{84}{\%} and \SI{71}{\%} at recalls of \SI{58}{\%} and \SI{60}{\%}, respectively, while RIO performs worst, with a precision of \SI{61}{\%} at \SI{51}{\%} recall (\Cref{tab:metrics}).
The reduced performance for the radar-based estimator is likely due to its generally larger high-frequency power content during healthy operation, which reduces the contrast between nominal and degraded behavior. This effect may be intrinsic to the estimator design but could also be influenced by labeling inaccuracies.
 Inaccuracies in our labeling approach are most pronounced at low velocities, which often coincide with the onset of estimator degradation. Exactly determining the moment when an estimate transitions from healthy to faulty is inherently ambiguous. This uncertainty may result in mislabeled data near the boundary between reliable and degraded operation. 
\subsection{Threshold Selection Without Labeled Data}
\begin{figure}[h]
\centering
\includegraphics[width=\linewidth]{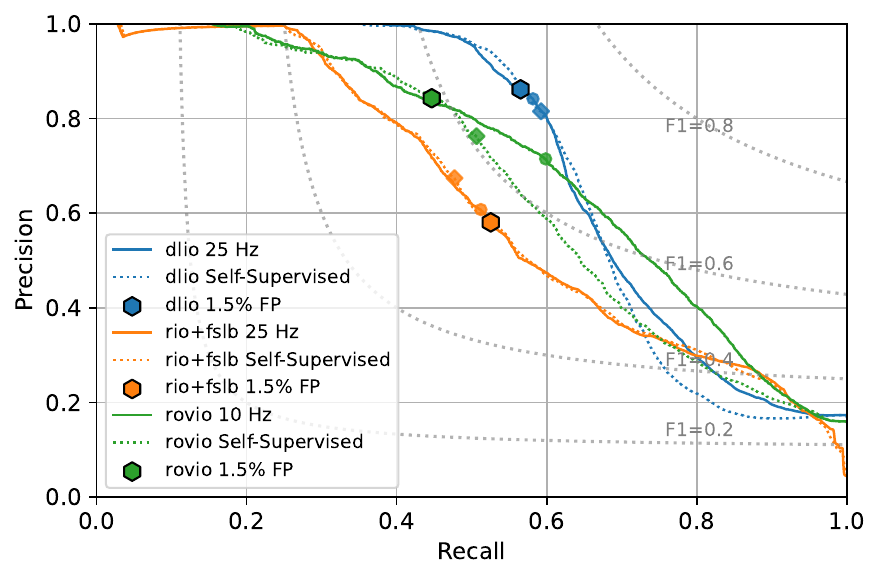}
\caption{Precision-Recall Curves for every estimator individually at the best performing and variable self-supervised cut-off frequency. The hexagonal marker represents the operating point which achieves maximal \SI{1.5}{\%} false positives. The other markers mark the thresholds that maximize the F1-score.}
\label{fig:pr_selected_sensors}
\end{figure}

In practice, it is often impractical to collect a large number of datasets featuring degraded estimators, and then produce accurate labels for them. How can we find reasonable thresholds and cut-off frequencies without labels? To address this, we evaluate two heuristic strategies for determining the cut-off frequency using only healthy data (which is easy to get). The intuition behind our approach is that, looking at \Cref{fig:heatmap_mean_psd}, having only healthy data gives us only the top plots, and we hypothesize that the lack of power content at higher frequencies is filled only during failures. Therefore, we identify the frequencies where there is typically no power in normal, healthy estimates. We suggest two ways of doing this: the first is an absolute threshold, where we select the highest frequency band whose power does not exceed $10^{-5}$\,\si{\metre\squared\per\second\squared}. Due to the velocity-dependence of the estimates, we split the velocity range into bins of \SI[per-mode=symbol]{0.1}{\metre\per\second}, and compute a cut-off per bin. The second method uses a relative threshold instead, selecting the cut-off frequency above which the power does not exceed \SI{0.5}{\%} of the median total power at that velocity.
While the fixed power-based cut-off performs poorly, the relative cut-off achieves comparable performance across the combined estimator set and for individual estimators (\Cref{fig:pr_selected_sensors}).
Interestingly, setting the power threshold such that only \SI{1.5}{\%} of healthy data exceed it, effectively corresponding to a \SI{1.5}{\%} false positive rate, and combining this threshold with the relative cut-off frequency yields performance close to that of estimator-specific tuning, without requiring labeled data of failures. Under this configuration, our method achieves the best overall performance on DLIO, with a precision of \SI{86}{\%} at a recall of \SI{56.5}{\%}. On RoVIO, it attains similarly high precision but at a substantially lower recall, falling below \SI{50}{\%}. This reduced recall is likely due to RoVIO’s tendency to degrade slowly and gradually rather than fail abruptly, as illustrated in \Cref{fig:covariances}. Such smooth drift results in a slow but steady increase in estimated velocity that can be difficult to distinguish from legitimate motion, particularly in the absence of pronounced high-frequency disturbances. This behavior is further reflected in the low optimal threshold for RoVIO—nearly two orders of magnitude smaller than for the other estimators. While effective on the evaluated dataset, such a low threshold is likely to increase false positives, and thus reduce precision, in scenarios with more uniformly distributed or higher-velocity motion.
\begin{table}[h]
\centering
\caption{Metrics for different state estimators at various cut-off frequencies. Shaded rows indicate the best PR-AUC per estimator. Italicized rows indicate that only healthy data was used to set $f_{co}$ and threshold. }
\label{tab:metrics}
\begin{tabular}{
l
r
r
r
r
r
}
\toprule
State Est. & $f_\text{co}$ & $th_{f1}$ & PR-AUC & Precision & Recall \\
           & {[Hz]}       & {$(m/s)^2$} & {[-]} & {[-]} & {[-]} \\
\midrule
All & $10^{-5}$ & 5.61e-4 & 0.58 & 0.785 & 0.437 \\
All & $0.5\%$ & 8.14e-4 & 0.61 & 0.769 & 0.484 \\
\rowcolor{gray!10}
All & 4 & 1.293e-3 & 0.636 & 0.782 & 0.497 \\
All & 10 & 7.11e-4 & 0.59 & 0.806 & 0.45 \\
All & 12 & 6.23e-4 & 0.58 & 0.809 & 0.435 \\
\midrule
\rowcolor{gray!10}
DLIO+RIO & 25 & 2.08e-4 & 0.684 & 0.783 & 0.546 \\
DLIO+RIO & 30 & 1.66e-4 & 0.682 & 0.787 & 0.532 \\
DLIO+RIO & 45 & 6.10e-5 & 0.667 & 0.756 & 0.509 \\
\midrule
\rowcolor{gray!10}
DLIO & 25 & 2.25e-4 & 0.728 & 0.841 & 0.582 \\
DLIO & $0.5\%$ & 8.57e-4 & 0.716 & 0.815 & 0.592 \\
\textit{DLIO(1.5\%FP)} & $0.5\%$ & 9.59e-4 & 0.716 & 0.861 & 0.565 \\
\midrule
\rowcolor{gray!10}
RIO & 25 & 1.35e-4 & 0.627 & 0.607 & 0.512 \\
RIO & $0.5\%$ & 7.17e-4 & 0.628 & 0.674 & 0.477 \\
\textit{RIO(1.5\%FP)} & \textit{0.5}\si{\%} & 5.07e-4 & 0.628 & 0.581 & 0.525 \\
\midrule
\rowcolor{gray!10}
RoVIO & 10 & 5.00e-6 & 0.708 & 0.715 & 0.599 \\
RoVIO & $0.5\%$ & 2.40e-5 & 0.664 & 0.763 & 0.507 \\
\textit{RoVIO(1.5\%FP)} & $0.5\%$ & 3.60e-5 & 0.664 & 0.843 & 0.447 \\
\bottomrule
\end{tabular}
\end{table}

\subsection{Performance On Groundtruth Dataset}
\begin{figure}[h]
\centering
\includegraphics[width=\linewidth]{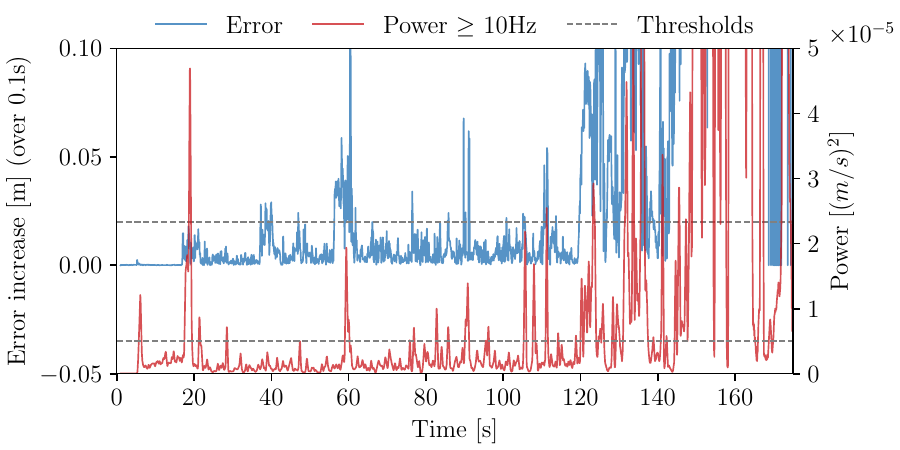}
\caption{Comparison of high-frequency power content and relative pose error (RPE) for RoVIO on the aerial dataset. Shown are the RPE over \SI{0.1}{\second} windows (blue), the velocity power content above \SI{10}{\hertz} (red), and their respective thresholds (dashed). Increases in high-frequency power coincide with estimator degradation and divergence, notably around \SI{60}{\second} and \SI{120}{\second}. Early power spikes occur during takeoff despite low RPE, likely due to unconverged IMU biases.}
\label{fig:metric_vs_rpe}
\end{figure}
To further showcase the introspectiveness of our proposed approach we evaluate its performance on an aerial dataset that includes state estimates from RoVIO and ground truth poses (also discussed in \cref{sec:covariances}). Our approach quantifies the degradation of estimates and is not a measure of accuracy. We therefore compare the high frequency power content to the relative pose error (RPE) between RoVIO and the ground truth. To classify the data we again divide the estimates into windows of \SI{0.1}{\second}. For every window we compute the RPE and again label estimates as faulty if the RPE exceeds \SI{0.2}{\metre}.
\Cref{fig:metric_vs_rpe} shows the power content RoVIO's velocity estimate for $f_{co} = 10$ \si{\hertz}, the threshold limit at $5e^{-6}$, as well as the RPE with the corresponding threshold at \SI{0.02}{\metre}. Over the whole dataset, our method detects the estimate health correctly for \SI{85}{\%} of the data. Notably, at around \SI{120}{\second} when RoVIO diverges, the power content correctly increases. Also, at \SI{60}{\second} during which RoVIO drifts (see \Cref{fig:covariances}), we can observe a spike in high-frequency power content that exceeds the defined threshold.
Two instances in power increase can be seen at the beginning of the dataset ($t \leq 20$\,\si{s}) even though the RPE remained low. During this time the platform was stationary and around \SI{20}{\second} took off. At that time, the \ac{IMU} biases have not converged yet, which most likely causes the increase in observed power. This indicates that the high-frequency power content does not only highlight degradation of the velocity estimate, but also when any velocity dependent state is inaccurately estimated. Additionally, inspecting the spectral characteristics of the \ac{IMU} could provide additional useful insight on estimator health.

\section{Results Summary}
\label{sec:results_summary}
The results presented show that the distribution of power content changes between reliable and unreliable estimates. We find that this behavior holds for estimators in general and that it depends neither on the fused sensor modality (camera, lidar, radar) nor on the used backend (filter-based or optimization-based). Finding a single cut-off frequency and power threshold for any state estimator is possible; our method achieves a PR-AUC of 0.64. However, our proposed method achieves better performance when these parameters are tuned individually for each state estimator (PR-AUC ranging from 0.63 to 0.73). Having labeled data helps set the $f_{co}$ and threshold, but we show that it is not strictly required. Using only healthy data, which is much easier to obtain, we achieve similar performance, as shown in \Cref{tab:metrics}. 
Additionally, we believe that the observed performance could likely be further improved by better labeled datasets. On one dataset for which ground truth was available, we correctly classified the estimator's condition for \SI{85}{\%} of the total estimates (see \Cref{sec:dataset}).
\section{Limitations}
\label{sec:limitations}
Frequency-domain analysis provides useful introspective indicators of estimator health, but still has several limitations that warrant discussion and motivate future work. Its effectiveness depends on increased high-frequency content: smooth or slowly accumulating drift, as well as complete sensor dropouts, often lack strong high-frequency signatures and are difficult to detect via spectral analysis, particularly at low velocities, where healthy and degraded signals are similar.

All estimators exhibit changes in spectral properties in the presence of degeneracies. However, design choices, such as filtering, optimization, and noise characteristics, can influence the amount of spectral content in higher frequency bands. This limits the generalizability of fixed thresholds and requires estimator-specific tuning to improve performance. Still, reasonable cut-offs and thresholds can often be estimated without hard-to-obtain failure-labeled data. More principled methods for automatic threshold selection could further improve robustness.

Finally, data collection and labeling remain significant challenges. Capturing diverse and representative failure modes requires in-the-wild operation, where ground truth is typically unavailable. Labeling via estimator consensus or manual inspection is imperfect, particularly near transition regions where degradation occurs gradually. Additionally, labeling is inherently application- and platform-dependent: in some tasks, millimeter-level errors are unacceptable, while in others, centimeter-scale drift is tolerable. This subjectivity directly affects labeling and evaluation. Our dataset is imbalanced, with failures occurring far less frequently than nominal operation, and biased toward low velocities, increasing the method's sensitivity to small changes in threshold selection. More diverse, application-specific datasets would enable more reliable threshold selection and help further characterize the method’s limitations.

\section{Conclusion} 
\label{sec:conclusion}
As robots are increasingly deployed in unstructured and unpredictable environments, the ability to assess the reliability of their own state estimates becomes critical. Unlike humans, robotic systems typically lack introspection and may continue operating on faulty estimates, where even small errors can have severe consequences.
This paper introduced a simple and widely applicable approach for detecting failures in state estimation. The method is independent of the specific sensor modality or estimation backend and treats the estimator as a black box, relying only on the fact that it performs IMU-based sensor fusion. At present, estimator-agnostic methods that assess the reliability of state estimates remain largely unexplored. In practice, the proposed method could inform downstream decisions and trigger safety mechanisms such as estimator resets, estimator switching, or sensor deactivation.

We showed that spectral analysis of estimated velocities provides an effective introspective signal. Estimator failures consistently manifest as increased high-frequency power in the power spectral density, while healthy estimates remain smooth. Intuitively, this behavior can be explained by internal disagreements in IMU-based sensor fusion, which lead to jumpy velocity estimates that are especially visible in higher frequency bands.
Experimental results show promising performance and, importantly, they reveal similar degradation patterns across different estimators and sensor modalities. The choice of cut-off frequency and detection threshold has a direct impact on the correct classification of estimator health. However, we demonstrated that suitable parameters can be found without requiring labeled failure data or ground truth. By tuning the cut-off frequency in a self-supervised manner and setting conservative thresholds, reliable detection can still be achieved even when only healthy data is available.
Overall, this work shows that frequency-domain analysis offers a practical and estimator-agnostic way to reason about state estimation reliability. Future work includes extending the approach to other robot platforms beyond aerial systems, incorporating additional signals such as IMU bias behavior, exploring velocity-dependent thresholds, and investigating learned variants of the method. Together, these directions could further strengthen introspective capabilities in real-world robotic systems.
\bibliographystyle{plainnat}
\bibliography{references}

\end{document}